\documentclass[sigconf,review=false,authordraft=false]{acmart}
\usepackage{booktabs}
\usepackage{threeparttable}
\usepackage{enumitem}
\usepackage[normalem]{ulem}
\usepackage{adjustbox}
\usepackage{array}
\usepackage{xparse}
\usepackage{amsmath}
\usepackage{graphicx}
\usepackage{float}
\acmConference[Conference acronym 'XX]{Make sure to enter the correct
  conference title from your rights confirmation email}{June 03--05,
  2018}{Woodstock, NY}
\acmYear{2018}
\copyrightyear{2018}
\AtBeginDocument{%
  }
\newcommand{\best}[1]{\textbf{#1}}
\newcommand{\second}[1]{\underline{#1}}

\newsavebox{\RAHCTableBox}
\NewDocumentEnvironment{fullwidthtabular}{m +b}{%
  \sbox{\RAHCTableBox}{\begin{tabular}{#1}#2\end{tabular}}%
  \typeout{RAHC-TABLE natural=\the\wd\RAHCTableBox; target=\the\linewidth}%
  \ifdim\wd\RAHCTableBox>\linewidth
    \resizebox{\linewidth}{!}{\usebox{\RAHCTableBox}}%
  \else
    \begin{tabular*}{\linewidth}{#1}#2\end{tabular*}%
  \fi
}{}

\setcopyright{none}
\renewcommand{\footnotetextcopyrightpermission}[1]{}

\title{RAHC: Rule-Augmented Hyper-Relational Knowledge Graph Completion for Materials Synthesis}

\author{Jiawei Sun}
\affiliation{%
  \institution{University of New South Wales}
  \city{Sydney}
  \country{Australia}
}
\email{jiawei.sun2@unsw.edu.au}

\author{Sijia Xu}
\affiliation{%
  \institution{University of New South Wales}
  \city{Sydney}
  \country{Australia}
}
\email{sijia.xu@unsw.edu.au}

\author{Zihan Chen}
\affiliation{%
  \institution{University of New South Wales}
  \city{Sydney}
  \country{Australia}
}
\email{zihan.chen6@unsw.edu.au}

\author{Zhengyi Yang}
\affiliation{%
  \institution{The University of Sydney}
  \city{Sydney}
  \country{Australia}
}
\email{zhengyi.yang@sydney.edu.au}

\renewcommand{\shortauthors}{Sun et al.}

\makeatletter
\AtBeginDocument{%
  \@ACM@reviewfalse
  \let\ACM@linecountL\@empty
  \let\ACM@linecountR\@empty
}
\makeatother

\begin{document}
\raggedbottom

\begin{abstract}
Scientific discovery relies on iterative experimentation and analysis, with experimental knowledge scattered across research papers and online repositories. However, experimental records are often incomplete, and recovering missing information requires domain knowledge and accumulated expertise. For example, in materials synthesis, inferring missing conditions requires interpreting the target material and experimental context using established synthesis knowledge. Hyper-relational knowledge graphs (HKGs) represent such knowledge through primary relations enriched with contextual qualifiers. Existing HKG completion methods learn structural representations, but domain knowledge and recurring experimental regularities remain implicit, limiting their direct use in reasoning about unseen objects. We propose RAHC, a rule-augmented HKG framework for materials synthesis condition completion, supporting single-slot prediction and joint multi-slot completion. RAHC constructs a traceable rule catalog from training records and retrieved literature, retrieves matching rules for query--candidate pairs, and separately encodes experimental structure and rule evidence. Hard rules constrain candidate validity, while soft rules guide ranking. RAHC-G aggregates evidence using global rule reliability, whereas RAHC-Q additionally accounts for query-specific relevance. Gated fusion combines rule evidence with structural representations. Experiments on five real-world materials synthesis datasets show the superiority of RAHC across single-slot and joint multi-slot completion tasks under Semi-inductive and Open-Answer Inductive settings, with Semi-inductive MRR improvements of up to 15.6\% over MAYPL. RAHC also achieves performance gains on the entity-disjoint ILPC'22 Small benchmark, demonstrating its generalization capability in the fully inductive setting. These results support experimental condition selection, synthesis planning, and the reuse of experimental knowledge in scientific discovery.
\end{abstract}
\maketitle

\section{Introduction}

Scientific discovery advances through hypothesis generation,
experimentation, and analysis of results~\cite{ghareeb2026multi,wang2023scientific}.
Knowledge accumulated from prior experiments informs subsequent
hypotheses, the selection of experimental conditions, and experiment
planning~\cite{king2004functional}. Experimental records carry this
knowledge by documenting the objects studied, the conditions used,
and the resulting observations. However, experimental knowledge is
scattered across scientific publications and online repositories,
making it difficult to consolidate and reuse~\cite{kononova2019dataset}.
Even when relevant records are available, they may lack key
experimental information because details are omitted, routine
procedures abbreviated, or domain knowledge left
implicit~\cite{davies2019digitization}. An important problem in reasoning with scientific knowledge is
therefore how to combine observed experimental context with
accumulated knowledge to infer and complete missing information in
experimental records, enabling fuller reuse of experimental
knowledge in subsequent research.

\begin{figure}[t]
    \centering
    \raisebox{0pt}[0.587408313\columnwidth][0pt]{%
        \includegraphics[width=\columnwidth,height=0.587408313\columnwidth,keepaspectratio]{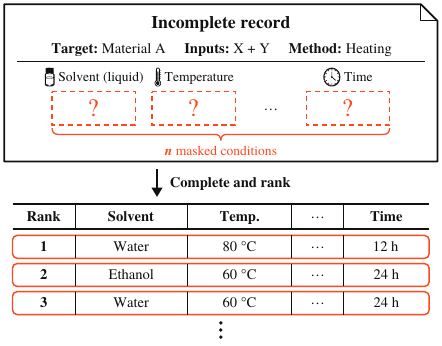}%
    }
    \caption{Joint completion of $n$ masked conditions. Candidate values and ranking are illustrative.}
    \Description{An incomplete experiment record lists known target material, inputs, and method, with n masked conditions illustrated by solvent, temperature, an ellipsis, and time. An arrow leads to ranked completions, where each row supplies one complete candidate combination for the missing fields. All values and ranks are schematic.}
    \label{fig:masking-tasks}
\end{figure}

Materials synthesis provides a representative setting for studying
this problem. A synthesis record connects a target material with the
precursors, solvent, temperature, reaction time, and synthesis method
used in a particular experiment; together, these describe a specific
synthesis route. For example, a record may specify the target material,
precursors, and temperature while omitting the solvent or reaction
time. Text mining and large language models can extract and structure
experimental information reported in scientific
publications~\cite{kononova2019dataset,dagdelen2024extraction}, but
records that remain incomplete after this process require inference
using established synthesis knowledge and the current experimental
context~\cite{vaucher2020automated}. Because the same material may be
prepared through different routes, the target material alone does
not determine a missing condition. Prediction must also consider the
other conditions observed in that experiment to assess whether a
candidate fits the current synthesis route. When several conditions
are missing, their dependencies must also be considered to assess
whether candidate conditions form a compatible combination. We define
this task as \textbf{context-aware synthesis condition completion}:
given a target material and the conditions observed in the same
experiment, use established synthesis knowledge to rank candidate
values for a missing condition or candidate combinations for multiple
missing conditions (Figure~\ref{fig:masking-tasks}).

A natural approach to these tasks is to represent experimental
records as relational facts and infer missing conditions through
knowledge graph completion.
KG and n-ary models learn relational patterns, while HKG models
additionally encode contextual qualifiers, providing a basis for
representing the conditions associated with an
experiment~\cite{lee2023ingram,zhu2021nbfnet,yin2025nshart,galkin2020stare,lee2025maypl}.
However, learned representations leave the experimental
regularities underlying a prediction implicit, making it
difficult to identify explicit support for a candidate under
the observed conditions.
Rule-based methods expose supporting
premises~\cite{meilicke2019anyburl,wang2025hyperpath},
but global confidence alone does not capture how relevant
a rule is to the current experimental context.
When several conditions are missing, an additional difficulty
arises from their interdependence.
Independent scoring does not account for interactions among
candidate values, while sequential completion can be sensitive
to the order in which conditions are filled. Overcoming these limitations requires addressing three key
challenges in context-aware synthesis condition completion.

\noindent\textit{\uline{C1: Connecting experimental context with explicit evidence.}}
Identifying explicit support for a candidate requires retrieving
relevant synthesis knowledge and determining whether its premises
hold under the observed experimental conditions.

\noindent\textit{\uline{C2: Distinguishing rule reliability from query relevance.}}
Since globally reliable rules may offer limited support for a
particular query, rule evidence must be assessed for both overall
reliability and relevance to the observed context and candidate.

\noindent\textit{\uline{C3: Evaluating dependencies among missing conditions.}}
Joint completion requires assessing the compatibility of candidate
conditions while distinguishing violations of explicit constraints
from combinations with weaker empirical support.

To this end, we propose \textbf{RAHC}, a
\underline{\textbf{R}}ule-\underline{\textbf{A}}ugmented
\underline{\textbf{H}}yper-relational knowledge graph
\underline{\textbf{C}}ompletion framework for
context-aware synthesis condition completion.
RAHC represents each synthesis experiment as a
Material--SynthesisRun fact enriched with contextual qualifiers
and retrieves rules derived from training records and literature
to provide explicit evidence for each query--candidate pair (\textit{C1}).
To distinguish global reliability from query-specific relevance,
RAHC-G weights matched rules by the former, whereas RAHC-Q
additionally learns the latter; a candidate-specific gate then
integrates rule evidence with structural representations (\textit{C2}).
For multiple missing conditions, rules with tuple conclusions
provide cross-slot evidence, while hard rules filter candidates
that violate explicit constraints and soft rules guide the
ranking of the remaining combinations (\textit{C3}).
We evaluate RAHC on single-slot and joint multi-slot completion
across five materials synthesis datasets, and further assess
its use of relational rule evidence on the entity-disjoint
ILPC'22 Small benchmark.
Our contributions are as follows:
\begin{itemize}[wide=0pt, leftmargin=*]
    \item We propose RAHC, a rule-augmented HKG framework for
    context-aware synthesis condition completion, which preserves
    the context of individual experiments and constructs a
    traceable rule catalog from training records and retrieved
    literature.

    \item We develop two complementary mechanisms for integrating
    rule evidence: RAHC-G aggregates matched rules according to
    global reliability, while RAHC-Q additionally accounts for
    query-specific relevance. A candidate-specific gate fuses
    the resulting evidence with structural representations.

    \item We extend single-slot prediction to joint multi-slot completion by combining individual condition scores with rules over condition combinations. Hard constraints exclude invalid candidates, while soft evidence guides the ranking of valid candidates.

    \item Experiments on five materials synthesis datasets
    demonstrate the effectiveness of RAHC. On MOF-SynHKG, the
    best-performing variants improve Semi-inductive MRR over
    MAYPL by 14.8\%, 15.6\%, and 6.4\% for solvent, temperature,
    and time completion, respectively. Additional gains on
    the entity-disjoint ILPC'22 Small benchmark support its
    applicability to inductive KG completion beyond materials
    synthesis.
\end{itemize}

\section{Related Work}

\subsection{Scientific Knowledge Graphs}

Synthesis records describe materials, participants, conditions, and sources. Completing a condition first requires a graph representation that preserves which experiment it belongs to. Triple-based scoring uses translations in TransE, complex-valued interactions in ComplEx, and rotations in RotatE~\cite{bordes2013transe,trouillon2016complex,sun2019rotate}; R-GCN and CompGCN aggregate relational neighborhoods~\cite{schlichtkrull2018rgcn,vashishth2020compgcn}. Once separate synthesis routes have been merged, however, a compatibility score cannot recover a condition's run identity.

Material KGs use these relations to integrate fragmented scientific records. MatKG connects literature-derived materials, properties, applications, and synthesis methods with source text and publication identifiers~\cite{venugopal2024matkg}. MOF-KG integrates synthesis, structure, composition, and publication knowledge~\cite{an2022mofkg}. This supports retrieval; predicting a missing value additionally requires aligning source evidence with one experiment, its observed conditions, and the proposed candidate.

To populate such graphs with experimental conditions, text-mining pipelines recover inorganic procedures~\cite{kononova2019dataset,wang2022solution}, SynMOF and ZeoSyn organize MOF and zeolite conditions~\cite{luo2022synmof,pan2024zeosyn}, and LLMs extract scientific entities and relationships~\cite{dagdelen2024extraction}. Extraction establishes what was reported; completion ranks values absent from the current record using its context and regularities across records.

Because one experiment contains interacting roles and conditions, higher-arity models retain multiple participants: NaLP uses role--value sets, HypE incorporates positions, and GRAN applies within-fact graph attention~\cite{guan2019nalp,fatemi2020hype,wang2021gran}. HINGE distinguishes an HKG's primary triple from its qualifiers~\cite{rosso2020hinge}. This lets a material--run fact retain its conditions as context. New materials and experiments then require inference from previously observed structure.

\subsection{Inductive HKG Completion}

Even when a material--run fact retains conditions, missing-value prediction must use its observed qualifiers: a solvent may have different relevance under different precursors, methods, or temperatures. StarE encodes qualifiers in messages, Hy-Transformer uses contextual auxiliary training, and HAHE combines global and local attention~\cite{galkin2020stare,yu2021hytransformer,luo2023hahe}. ShrinkE models qualifier monotonicity, implication, and exclusion geometrically~\cite{xiong2023shrinke}. These models learn contextual compatibility, but their scores do not identify the documented regularity, evidence, or scope supporting a particular candidate.

Synthesis research continually introduces materials and runs absent from training, making entity-specific embeddings insufficient for these queries. Inductive methods instead use subgraphs and paths in GraIL and NBFNet, compositional tokens in NodePiece, or relation graphs in InGram~\cite{teru2020grail,zhu2021nbfnet,galkin2022nodepiece,lee2023ingram}. Qualifier-aware prediction, NS-HART, and MAYPL extend induction to richer facts~\cite{ali2021inductive,yin2025nshart,lee2025maypl}. Representing an unseen object still differs from identifying the experimental regularities supporting its candidate conditions.

An unseen material may share chemicals and condition values with training records, whereas an inference graph may contain entirely disjoint entities. Benchmarks distinguish these inductive settings~\cite{kochsiek2023benchmark,galkin2022ilpc}. We assess unseen material queries and separately test entity-disjoint transfer on ILPC'22. In synthesis, support must also be tied to the particular run and its observed conditions, not just to the unseen material. Structural induction alone does not identify why a candidate is supported, motivating explicit rules.
\begin{figure}[t]
\centering
\includegraphics[width=0.9\columnwidth]{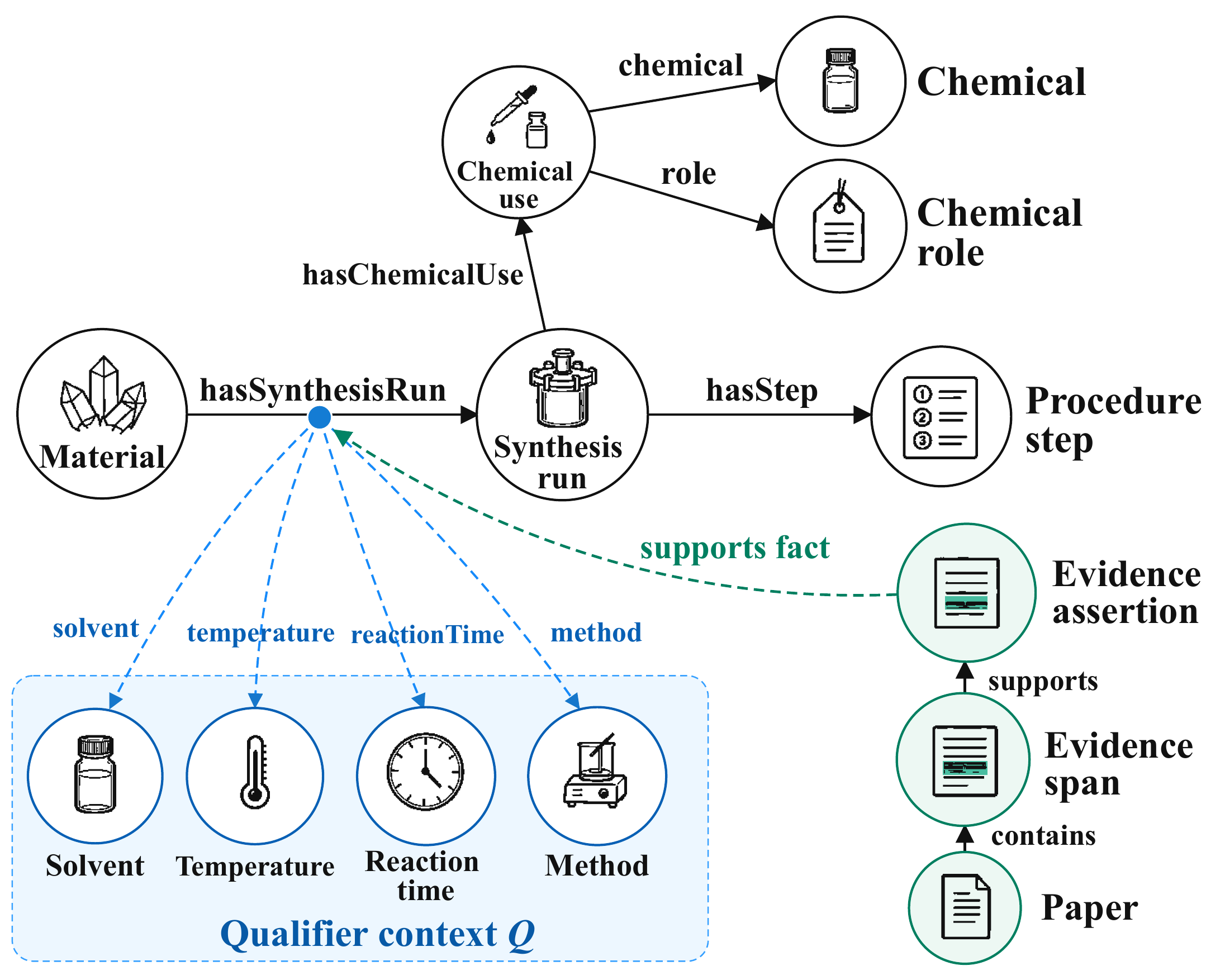}
\caption{Run-centric synthesis HKG. Qualifiers contextualize the material--run fact; chemical-use roles, procedure steps, and provenance retain experimental semantics.}
\Description{A Material node connects to SynthesisRun through hasSynthesisRun. The primary fact has contextual qualifier links to solvent, temperature, reaction time, and method. SynthesisRun also connects to chemical-use and procedure-step nodes. ChemicalUse distinguishes Chemical from ChemicalRole. Paper, EvidenceSpan, and EvidenceAssertion form a provenance chain supporting the fact.}
\label{fig:synthesis-hkg-schema}
\end{figure}
\subsection{Rule-Augmented Reasoning}

Condition ranking needs to identify which prior experimental pattern supports which candidate. Rules expose premises and conclusions: AMIE+ and AnyBURL mine Horn rules, Neural LP and DRUM make learning differentiable, and RNNLogic jointly trains generation and prediction~\cite{galarraga2015amieplus,meilicke2019anyburl,yang2017neurallp,sadeghian2019drum,qu2021rnnlogic}. Such evidence is inspectable, but predictive coverage depends on rule groundings; aggregate confidence alone does not establish relevance to the current material, method, and remaining conditions.

Rule coverage can be sparse for a new experiment, so symbolic evidence must work alongside structural representations. IterE alternates embedding learning and axiom induction; RulE jointly embeds entities, relations, and rules~\cite{zhang2019itere,tang2024rule}. HyperPath incorporates qualifier-rich facts and paths~\cite{wang2025hyperpath}. For synthesis, historical reliability and current-query relevance must be distinguished, and a rule's conclusion must remain aligned with the value or tuple it supports.

When training records do not cover an emerging material, literature may supply additional experimental regularities. Retrieval-augmented generation grounds responses in documents~\cite{lewis2020rag}, while scientific extraction recovers structured statements~\cite{dagdelen2024extraction}. Document relevance, however, does not establish whether a regularity applies to the present run or supports the candidate being ranked. Rule evidence must connect its premises, scope, and conclusion to that candidate.

\begin{figure*}[t]
\centering
\includegraphics[width=\textwidth]{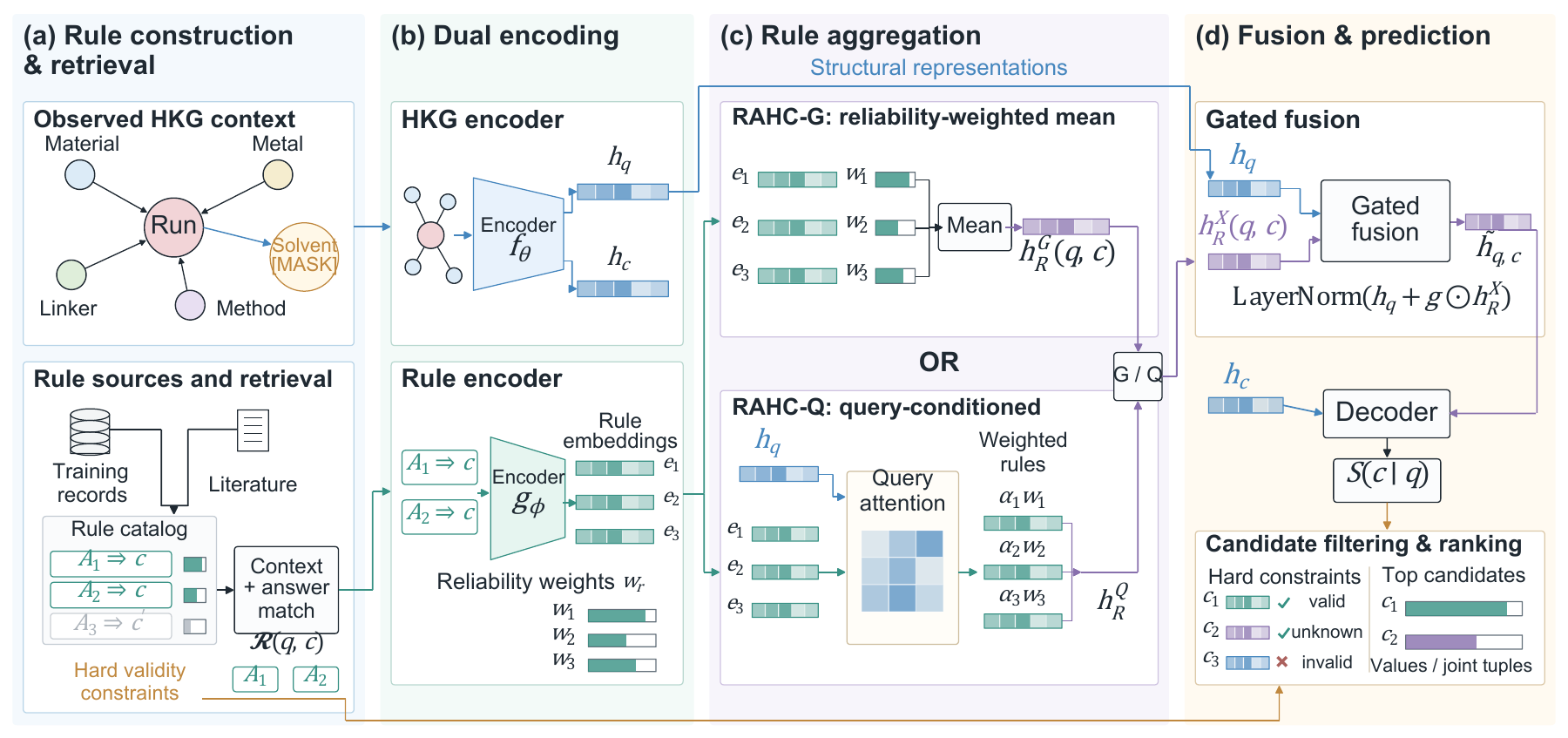}
\caption{The RAHC framework. (a) Training records and retrieved literature form a traceable rule catalog; retrieval matches observed context and candidate conclusions. (b) Separate encoders represent HKG structure and rule evidence. (c) RAHC-G aggregates by global reliability; RAHC-Q also learns query-conditioned relevance. (d) Gated fusion supports candidate-value and tuple ranking. Hard rules constrain candidate validity.}
\Description{Four modules show rule construction and candidate retrieval, parallel HKG and rule encoders, global or query-conditioned aggregation, and gated fusion with candidate ranking. Arrows distinguish structural representations, matched rule evidence, and validity constraints.}
\label{fig:rahc-framework}
\end{figure*}

\section{Problem Formulation}

\subsection{Hyper-Relational Graphs}

Let $\mathcal G=(\mathcal E,\mathcal R,\mathcal F)$ denote a knowledge graph, where $\mathcal E$, $\mathcal R$, and $\mathcal F$ are the entity, relation, and fact sets. A binary fact $(h,r,t)$ connects a head entity $h\in\mathcal E$ to a tail entity $t\in\mathcal E$ through $r\in\mathcal R$. We associate this primary triple with contextual relation--value pairs to define a hyper-relational fact~\cite{rosso2020hinge,galkin2020stare}:
\begin{equation}
\begin{aligned}
f&=(h,r,t,Q)\in\mathcal F,\\
Q&=\{(q_j,v_j)\}_{j=1}^{|Q|}\subseteq\mathcal R\times\mathcal E.
\end{aligned}
\end{equation}
Here $q_j$ is a qualifier relation and $v_j$ its value; $Q=\varnothing$ gives an ordinary triple. Qualifiers contextualize the entire primary triple. A type map $\tau:\mathcal E\to\mathcal T$ assigns entities to materials, synthesis runs, chemicals, chemical uses, procedure steps, condition values, or evidence in type set $\mathcal T$.

For illustration, let $m$ be one target material and $u_1,u_2$ two separate synthesis runs. Suppose the illustrative run $u_1$ uses Zn(NO$_3$)$_2$ and H$_2$BDC in DMF at $120\,^{\circ}\mathrm C$ for 24 h, while $u_2$ uses a different solvent and temperature. These are schematic values, not records claimed from the evaluated datasets. Both runs target $m$, but their qualifier sets remain separate; when predicting the solvent for $u_1$, the temperature of $u_2$ is not part of its observed context.

\subsection{Synthesis as Graph Facts}

For a target material $m$ and a synthesis run $u$, we define the primary synthesis fact as
\begin{equation}
f_u=(m,\texttt{hasSynthesisRun},u,Q_u).
\end{equation}
Run $u$ binds one experiment's inputs, conditions, and operations. Different routes to the same material have $u\neq u'$ and separate qualifier sets, even when values coincide. Each $(s,v)\in Q_u$ assigns value $v$ to a chemical role or condition $s$: metal source, linker, precursor, organic structure-directing agent (OSDA), solvent, temperature, reaction time, or method. Thus, solvent and temperature pairs within $Q_u$ describe the same experiment.

Supporting graph relations preserve information beyond these target slots (Figure~\ref{fig:synthesis-hkg-schema}). A chemical-use entity $x$ links run $u$ to a chemical $a$ and its run-specific role $\rho$ through the triple patterns
\begin{equation}
\begin{aligned}
&(u,\texttt{hasChemicalUse},x),\\
&(x,\texttt{chemical},a),\quad(x,\texttt{role},\rho).
\end{aligned}
\end{equation}
Amounts are attached to the chemical use, separating chemical identity, experimental role, and quantity. Likewise, the recorded operations of run $u$ form the set
\begin{equation}
\mathcal P_u=\{p\in\mathcal E:(u,\texttt{hasStep},p,\varnothing)\in\mathcal F\}.
\end{equation}
Each of the $|\mathcal P_u|$ steps retains its operation and conditions. To specify the evidence link in Figure~\ref{fig:synthesis-hkg-schema}, let $d,e,a$ be document, evidence-span, and assertion entities. The ordinary graph facts $(d,\texttt{contains},e,\varnothing)$ and $(e,\texttt{supports},a,\varnothing)$ identify the source chain. Because $f_u$ is a fact rather than an entity, assertion support is a fact-level provenance association
\begin{equation}
\mathcal L_{\rm prov}\subseteq\mathcal E_{\rm assert}\times\mathcal F,
\qquad(a,f_u)\in\mathcal L_{\rm prov}.
\end{equation}
This association corresponds to the ``supports fact'' link; the completion target remains a masked qualifier of $f_u$. Continuous values remain exact in the graph; ranking and rule-statistic bins are fitted on training data.

\subsection{Completion Queries}

We formulate synthesis-condition recovery as predicting missing qualifier values. A target \emph{slot} $s\in\mathcal R$ specifies the qualifier relation to complete. Let $\nu_s$ denote slot-specific value normalization, including training-fitted binning for numerical ranking targets. The training-only candidate vocabulary is
\begin{equation}
\mathcal C_s=\{\nu_s(v):(s,v)\in Q_u,\ f_u\in\mathcal F_{\rm tr}\},
\end{equation}
where $\mathcal F_{\rm tr}$ contains training facts. Candidate types follow the target slot: a solvent query ranks solvent values, while a temperature query ranks temperature bins. Amounts and procedural attributes remain available as context.

For fact $f_{u_i}$, let $M_i=\{s_{i1}^*,\ldots,s_{ik}^*\}$ be the set of masked slots. Removing every value of these relations gives the observed qualifier context
\begin{equation}
K_i=Q_{u_i}\setminus\{(s,v)\in Q_{u_i}:s\in M_i\}.
\end{equation}
The resulting query and its candidate domain are
\begin{equation}
\begin{aligned}
q_i^{(k)}&=(m_i,u_i,K_i,M_i),\\
\mathcal C_{M_i}&=\mathcal C_{s_{i1}^*}\times\cdots\times\mathcal C_{s_{ik}^*}.
\end{aligned}
\end{equation}
For $k=1$, this reduces to $q_i^{(1)}=(m_i,u_i,K_i,\{s_i^*\})$, and the model ranks $c\in\mathcal C_{s_i^*}$ by $S(c\mid q_i^{(1)})$. The recorded normalized values $\{\nu_{s_i^*}(v):(s_i^*,v)\in Q_{u_i}\}$ are the correct answers. A slot may contain multiple answers, such as mixed solvents, so all its values are masked together. In the schematic $u_1$, masking $(\texttt{solvent},\mathrm{DMF})$ leaves its material, precursors, $120\,^{\circ}\mathrm C$, 24 h, and method as context for ranking training-vocabulary solvents.

For $k\geq2$, the model scores each tuple $\mathbf c=(c_1,\ldots,c_k)\in\mathcal C_{M_i}$ by $S(\mathbf c\mid q_i^{(k)})$, inserting $(s_{ij}^*,c_j)$ into the same fact. Masking solvent and temperature in $u_1$ makes $(\mathrm{DMF},120\,^{\circ}\mathrm C)$ the schematic complete answer; pairing its solvent with $u_2$'s temperature is incorrect. Joint MRR requires the complete recorded answer, not a partial slot match.

\subsection{Evaluation Scenarios}
\label{sec:inductive-settings}

The materials-synthesis experiments use two primary protocols with unseen query objects. Write $\mathcal C_s^{\mathrm{tr}}=\mathcal C_s$ to emphasize the training-only candidate vocabulary. \textbf{Semi-inductive} evaluation includes queries whose correct answer lies in $\mathcal C_s^{\mathrm{tr}}$. \textbf{Open-Answer Inductive} evaluation also retains queries with correct answers outside $\mathcal C_s^{\mathrm{tr}}$ and scores those answers as prediction failures. The two protocols use the same training-only candidate vocabulary and differ in answer coverage.

We evaluate \textbf{Fully-inductive} transfer separately on ILPC'22 Small~\cite{galkin2022ilpc}. Let $\mathcal E_{\mathrm{tr}}$ and $\mathcal E_{\mathrm{inf}}$ be the entity sets of the training and inference graphs. Here $\mathcal E_{\mathrm{tr}}\cap\mathcal E_{\mathrm{inf}}=\varnothing$; head and tail entities are ranked over the inference-graph entities, with relation semantics shared across graphs.

\section{The RAHC Framework}
\label{sec:method}

For a masked query $q=(m,u,K,M)$, RAHC encodes HKG context, retrieves rules supporting each candidate $c$, and fuses their evidence before decoding $S(c\mid q)$ (Figure~\ref{fig:rahc-framework}). The rule branch uses either global reliability (RAHC-G) or query-conditioned attention (RAHC-Q). Both variants share the same candidate-scoring interface:
\begin{equation}
\begin{aligned}
(h_q,h_c)&=\operatorname{HKGEnc}(q,c,\mathcal G),\\
\widetilde h_{q,c}&=\operatorname{Fuse}(h_q,h_R^X(q,c)),\\
S(c\mid q)&=\operatorname{Dec}(\widetilde h_{q,c},h_c),
\end{aligned}
\end{equation}
where $X\in\{G,Q\}$. The structural representation $h_q$ depends on the observed run, whereas $h_R^X(q,c)$ contains rules matched to the proposed answer. This distinction lets different candidates for the same masked query receive different rule evidence. The interface also scores complete tuples when several slots are masked.

\subsection{Rule Construction and Retrieval}

A rule $r=(A_r,B_r,\Omega_r,E_r,CE_r,z_r,w_r)$ has antecedent conditions $A_r$, a candidate conclusion $B_r$, and an applicability scope $\Omega_r$. Supporting and counter-evidence $(E_r,CE_r)$ retain provenance; $z_r$ identifies the rule source and $w_r\in[0,1]$ its global reliability. Source, hard/soft strength, and single-/multi-slot conclusion are independent properties (Table~\ref{tab:rule-taxonomy}).

The catalog combines training-mined dependencies with literature-derived statements. Data rules use up to three generalizable antecedents; literature rules undergo structured extraction and reverse searches for support and counterexamples. Both predict values or tuples, exclude identifiers from antecedents, and use training-fitted numerical bins. Appendix~\ref{app:rule-construction} details construction and verification.

\paragraph{Validity and ranking evidence.}

Hard rules check target-slot types, elemental consistency, and mutually exclusive roles. Only $H(q,c)=\text{invalid}$ excludes $c$; valid or unknown candidates remain. Soft rules rank the survivors.

\paragraph{Global reliability.}

For data rules, support, confidence, and lift measure evidence volume, conditional frequency, and enrichment; significance, bootstrap stability, independent-source count, and counterexample rate provide further quality signals. Literature rules use independent support, textual directness, scope consistency, and counterexamples. Source-specific quality vectors are validation-calibrated and task-normalized into fixed $w_r\in[0,1]$. All mining and rule statistics use training records only; Appendix~\ref{app:rule-construction} gives the definitions.

\paragraph{Candidate-specific retrieval.}

Retrieval first matches the target slot, observed antecedents, and material/reaction scope. Among these query-matched rules $\mathcal R(q)$, candidate $c$ receives only $\mathcal R(q,c)=\{r\in\mathcal R(q):B_r=c\}$. A rule that predicts another solvent, for example, cannot support $c$ merely because its antecedent fits the run. Conversely, a rule with the right conclusion is inactive when its required conditions are absent or its scope excludes the query. If $\mathcal R(q,c)=\varnothing$, then $h_R^X(q,c)=\mathbf 0$ and the structural branch still predicts. Thus a rule-free candidate is not discarded unless a hard validity check explicitly rejects it.

\subsection{Dual Structure--Rule Encoding}

The HKG encoder captures the primary fact, observed qualifiers, and local graph, producing $(h_q,h_c)=\operatorname{HKGEnc}(q,c,\mathcal G)$. RuleEnc pools role--value embeddings $\phi(s,v)$ separately over $A_r$, $B_r$, and $\Omega_r$ into $a_r$, $b_r$, and $o_r$. With normalized evidence statistics $v_r$ and source embedding $z_r$, it computes
\begin{equation}
e_r=\operatorname{LayerNorm}\!\left(
\operatorname{MLP}([a_r;b_r;o_r;v_r;z_r])\right).
\end{equation}
DOIs, paper IDs, and run IDs are excluded from encoder features; $z_r$ encodes source type, not document identity. Projected structural and rule representations share a $d$-dimensional space, while evidence-derived $w_r$ remains separate from $v_r$.

\subsection{Reliability-Aware Rule Aggregation}

\paragraph{Global aggregation (RAHC-G)}

RAHC-G averages matched embeddings weighted by fixed reliability:
\begin{equation}
h_R^{G}(q,c)=\frac{1}{\max(1,|\mathcal R(q,c)|)}
\sum_{r\in\mathcal R(q,c)}w_r W_Ve_r.
\end{equation}
Here $W_V$ projects rule embeddings; retrieval depends on $(q,c)$, but $w_r$ does not.

\paragraph{Query-conditioned aggregation (RAHC-Q)}

RAHC-Q instead uses $h_q$ to attend to matched rules, with $w_r$ as a reliability prior:
\begin{equation}
\begin{aligned}
\alpha_{q,c,r}&=\operatorname{softmax}_{r\in\mathcal R(q,c)}(a_{q,r}),\\
a_{q,r}&=\frac{(W_Qh_q)^\top(W_Ke_r)}{\sqrt d}
+\eta\log(w_r+\epsilon).
\end{aligned}
\end{equation}
Here $W_Q,W_K$ measure query--rule compatibility, $\eta$ controls the prior, and $\epsilon$ stabilizes the logarithm. Attention is normalized within $\mathcal R(q,c)$, giving
\begin{equation}
h_R^{Q}(q,c)=\sum_{r\in\mathcal R(q,c)}\alpha_{q,c,r}w_rW_Ve_r.
\end{equation}
The two variants use the same matched rule set but make different weighting decisions. RAHC-G uses each rule's calibrated reliability across all queries; RAHC-Q can emphasize a different subset for the current run through $\alpha_{q,c,r}$. Neither attention nor reliability changes which candidate a retrieved rule concludes. This separates evidence strength, contextual relevance, and candidate-specific retrieval.

\subsection{Fusion and Joint Prediction}

For $X\in\{G,Q\}$, a candidate-specific gate adds rule evidence to the structural query:
\begin{equation}
\begin{aligned}
g_{q,c}&=\sigma(W_g[h_q;h_R^X(q,c)]),\\
\widetilde h_{q,c}&=\operatorname{LayerNorm}
\bigl(h_q+g_{q,c}\odot h_R^X(q,c)\bigr).
\end{aligned}
\end{equation}
Here $[\cdot;\cdot]$ concatenates, $\sigma$ is sigmoid, and $\odot$ is element-wise multiplication. A decoder scores $(\widetilde h_{q,c},h_c)$.

For jointly masked slots $M=\{s_1^*,\ldots,s_k^*\}$ and a complete candidate tuple $\mathbf c\in\mathcal C_M$, define $C(\mathbf c)=\{(s_j^*,c_j)\}_{j=1}^k$. Joint evidence consists of query-matched rules whose consequents agree with at least two target slots:
\begin{equation}
\begin{split}
\mathcal R_J(q,\mathbf c)=\{r\in\mathcal R(q):\;&B_r\subseteq C(\mathbf c),\\
&|\operatorname{slots}(B_r)|\geq 2\}.
\end{split}
\end{equation}
Here $\operatorname{slots}(B_r)$ denotes consequent roles. A solvent--temperature rule can support a larger tuple when both of its conclusions agree with that tuple, but it cannot reward a combination with a different temperature. Single-slot rules are excluded from $\mathcal R_J$ because their evidence already enters the component scores. This prevents counting the same rule twice while allowing explicit support for cross-condition combinations. G/Q aggregation over $\mathcal R_J$ yields $h_{R,J}^{X}(q,\mathbf c)$. For tuple embedding $h_{\mathbf c}$, define
\begin{equation}
\begin{gathered}
\ell_J^X=v_J^\top\tanh\!\left(
W_qh_q+W_ch_{\mathbf c}+W_Rh_{R,J}^{X}\right)+b_J,\\
S_R^X(\mathbf c\mid q)=
\begin{cases}
0,&\mathcal R_J(q,\mathbf c)=\varnothing,\\
\operatorname{softplus}(\ell_J^X),&\text{otherwise}.
\end{cases}
\end{gathered}
\end{equation}
The learned projections, $v_J$, and $b_J$ give a nonnegative rule score that is zero without joint evidence. The full tuple score is
\begin{equation}
\begin{aligned}
S(\mathbf c\mid q^{(k)})={}&
\sum_{j=1}^{k}S(c_j\mid q^{(k)},s_j^*)\\
&+\lambda_J S_R^X(\mathbf c\mid q^{(k)}).
\end{aligned}
\end{equation}
All component scores use the same jointly masked query; $\lambda_J$ controls joint evidence. Without it, the score is the sum of slot scores. Evaluation requires a complete-tuple match.

\paragraph{Learning.}

Candidate cross-entropy trains single-answer queries, with correct-answer probability mass aggregated for multi-answer queries and the same objective applied to tuples. Training updates the encoders, fusion, decoder, and RAHC-Q attention; global reliabilities stay fixed. Appendix~\ref{app:rule-construction} gives the objective.

Splits precede candidate construction and rule mining. Validation selects calibration, hyperparameters, and checkpoints. Masking removes every target-slot value, equivalent alias path, and answer-revealing derived field. Identifiers serve only partitioning, source counts, and traceability, never rule antecedents; test answers do not enter the rule catalog.

\section{Experiments}
\label{sec:experiments}

Our experiments test three questions: whether explicit rules improve condition ranking within materials synthesis, whether the same mechanism transfers across material families and target slots, and whether it remains useful when training and inference entities are disjoint. We use MOF synthesis as the primary case because a run combines metal sources, linkers, solvents, temperature, time, and procedural context; this mixture of roles and conditions makes it a natural test of run-centric HKG representation and candidate-specific rule evidence~\cite{luo2022synmof}. We then evaluate inorganic-precursor and zeolite-OSDA completion to test transfer beyond MOFs, and use ILPC'22 Small to isolate fully inductive KG generalization. The setup below introduces the datasets, protocols, comparators, and metrics before the task-specific results.

\subsection{Experimental Setup}

Our code is available at \url{https://anonymous.4open.science/r/RAHC-2B48}.

\paragraph{Datasets.}

We use five materials-synthesis datasets. MOF-SynHKG is the main HKG for solvent, temperature, and reaction-time completion, including joint completion; SynMOF and FAIR-MOFs evaluate the same condition types across MOF data sources~\cite{luo2022synmof}. Solution-based Inorganic evaluates missing-precursor completion, while ZeoSyn evaluates organic structure-directing agent (OSDA) completion in zeolite synthesis~\cite{wang2022solution,pan2024zeosyn}. RAHC is retrained and its rules rebuilt for each dataset. Separately, ILPC'22 Small evaluates head and tail prediction with disjoint training and inference entities~\cite{galkin2022ilpc}. Further dataset and task information is provided in Appendix~\ref{app:dataset-overview}.

\begin{table*}[!t]
\centering
\caption{MOF solvent, temperature, and time completion (MRR). SI denotes Semi-inductive; OA denotes Open-Answer Inductive. Scores are rounded to three decimals; bold and underline mark the best and second-best displayed scores.}
\label{tab:mof-main}
\small
\setlength{\tabcolsep}{2.5pt}
\renewcommand{\arraystretch}{1.03}
\begin{adjustbox}{max width=\textwidth}
\begin{fullwidthtabular}{@{\extracolsep{\fill}}l *{15}{c}@{}}
\toprule
& \multicolumn{6}{c}{MOF-SynHKG} & \multicolumn{3}{c}{SynMOF} & \multicolumn{6}{c}{FAIR-MOFs} \\
\cmidrule(lr){2-7}\cmidrule(lr){8-10}\cmidrule(l){11-16}
& \multicolumn{2}{c}{Solvent} & \multicolumn{2}{c}{Temperature} & \multicolumn{2}{c}{Time} & Solvent & Temperature & Time & \multicolumn{2}{c}{Solvent} & \multicolumn{2}{c}{Temperature} & \multicolumn{2}{c}{Time} \\
\cmidrule(lr){2-3}\cmidrule(lr){4-5}\cmidrule(lr){6-7}\cmidrule(lr){8-8}\cmidrule(lr){9-9}\cmidrule(lr){10-10}\cmidrule(lr){11-12}\cmidrule(lr){13-14}\cmidrule(l){15-16}
Method & SI & OA & SI & OA & SI & OA & SI & SI & SI & SI & OA & SI & OA & SI & OA \\
\midrule
InGram & 0.013 & 0.012 & 0.543 & 0.286 & 0.458 & 0.134 & 0.157 & 0.517 & 0.451 & 0.021 & 0.020 & 0.534 & 0.530 & 0.456 & 0.450 \\
NBFNet & 0.319 & 0.303 & 0.675 & 0.356 & 0.629 & 0.184 & 0.520 & 0.661 & 0.625 & 0.478 & 0.467 & 0.848 & 0.842 & 0.549 & 0.542 \\
NS-HART & 0.315 & 0.300 & 0.551 & 0.290 & 0.603 & 0.176 & 0.432 & 0.614 & 0.603 & 0.436 & 0.426 & 0.834 & 0.828 & 0.554 & 0.547 \\
HYPER & 0.321 & 0.305 & 0.669 & 0.353 & 0.636 & 0.186 & 0.660 & 0.637 & 0.621 & 0.461 & 0.450 & 0.868 & 0.862 & 0.572 & 0.565 \\
QBLP & 0.322 & 0.306 & 0.665 & 0.350 & 0.644 & 0.188 & 0.684 & 0.648 & 0.623 & 0.442 & 0.431 & 0.872 & 0.866 & 0.542 & 0.535 \\
MAYPL & 0.330 & 0.314 & 0.633 & 0.334 & 0.640 & 0.187 & 0.662 & 0.555 & 0.438 & 0.416 & 0.406 & 0.849 & 0.843 & 0.523 & 0.516 \\
AnyBURL & \second{0.370} & \second{0.352} & 0.704 & 0.371 & 0.661 & 0.193 & 0.701 & 0.700 & 0.597 & \second{0.545} & \second{0.532} & 0.824 & 0.818 & 0.578 & 0.570 \\
HyperPath & 0.364 & 0.346 & 0.720 & 0.379 & 0.670 & 0.196 & \second{0.741} & 0.698 & 0.626 & 0.533 & 0.520 & \second{0.875} & \second{0.869} & 0.573 & 0.566 \\
RAHC-G & 0.369 & 0.351 & \second{0.731} & \second{0.385} & \best{0.681} & \best{0.199} & \best{0.745} & \second{0.705} & \second{0.628} & \best{0.552} & \best{0.539} & 0.873 & 0.867 & \second{0.583} & \second{0.575} \\
RAHC-Q & \best{0.379} & \best{0.360} & \best{0.732} & \best{0.386} & \second{0.673} & \second{0.197} & 0.732 & \best{0.713} & \best{0.634} & 0.538 & 0.525 & \best{0.878} & \best{0.871} & \best{0.591} & \best{0.583} \\
\bottomrule
\end{fullwidthtabular}
\end{adjustbox}
\end{table*}

\paragraph{Evaluation protocols.}

We evaluate material-condition queries under two answer-coverage scenarios defined in Section~\ref{sec:inductive-settings}. Both rank the same slot-typed candidate vocabulary constructed from training facts, so test answers never expand the candidate set. Semi-inductive (SI) includes unseen query objects whose recorded answer is in that vocabulary. Open-Answer Inductive (OA) retains the additional queries whose recorded answer is outside it and counts those as failures. Thus SI measures ranking among representable answers, whereas OA also measures the effect of answer coverage. We split records before constructing candidates or mining rules, derive rule statistics only from training records, and mask every true value of a target slot within a test run to prevent answer leakage.

For standard Fully-inductive evaluation, we use ILPC'22 Small~\cite{galkin2022ilpc}: its training and inference graphs have disjoint entity sets. Observed inference-graph edges provide context for 6,653 unseen entities, while validation and test target edges are withheld. We rank both heads and tails against all inference entities with filtered ranking and realistic tie handling; graph sizes are reported in Appendix~\ref{app:ilpc22-details}.

On ILPC'22, both RAHC variants share the NodePieceGNN structural encoder~\cite{galkin2022nodepiece,galkin2022ilpc} and match training-mined relation rules to observed inference-graph paths. NodePiece, NodePieceGNN, NBFNet, and AnyBURL use the same queries and candidates. Validation MRR selects seeds, checkpoints, and fusion coefficients before one test evaluation per method.

\paragraph{Baselines.}

We select comparators to separate the contributions of structural generalization, qualifier-aware modeling, and explicit rules. InGram~\cite{lee2023ingram} and NBFNet~\cite{zhu2021nbfnet} test inductive relation and path reasoning; NS-HART~\cite{yin2025nshart}, MAYPL~\cite{lee2025maypl}, and the HYPER and QBLP adaptations test higher-arity or hyper-relational structure; AnyBURL~\cite{meilicke2019anyburl} and HyperPath~\cite{wang2025hyperpath} test symbolic rule and path evidence. Transductive comparisons additionally include StarE and HAHE. Methods share task-specific queries, candidates, and evaluation procedures within each result table. On ILPC'22, NodePiece and NodePieceGNN provide benchmark-compatible structural comparisons alongside NBFNet and AnyBURL. For joint completion, \emph{Independent} is the same RAHC tuple scorer with $\lambda_J=0$, summing its component slot scores; ordered and bidirectional autoregressive baselines fill masked slots sequentially.

\paragraph{Metrics.}

The primary metrics are mean reciprocal rank (MRR) and Hits@$1$, Hits@$3$, and Hits@$10$. Materials tables report MRR under the Semi-inductive and Open-Answer Inductive protocols. Temperature and time tasks also use macro-F1 and absolute error based on fixed representative values. ILPC'22 additionally reports Hits@$5$ and adjusted mean rank index (AMRI); its paired bootstrap resamples test triples after averaging head- and tail-prediction reciprocal ranks.

\subsection{Single-Slot MOF Completion}
\label{sec:single-slot}

We first test whether rules improve individual condition rankings across three MOF data sources. Table~\ref{tab:mof-main} reports solvent, temperature, and time completion under the SI and OA protocols. Within a directly comparable column, the best result is \textbf{bold} and the second best is \underline{underlined}.

\paragraph{Solvent prediction.}
A RAHC variant ranks first in all five solvent columns. On MOF-SynHKG, RAHC-Q reaches SI and OA MRR of 0.379 and 0.360, exceeding MAYPL by 14.8\% and 14.6\%, respectively. RAHC-G leads SynMOF at 0.745 and FAIR-MOFs at 0.552 SI and 0.539 OA MRR. AnyBURL and HyperPath are also competitive, placing explicit rule evidence close to the leading structural methods or above them. Combining that evidence with structural representations yields the strongest solvent ranking in each source, although the preferred aggregation strategy varies.

\paragraph{Temperature and time prediction.}
A RAHC variant leads all ten temperature and time columns. On MOF-SynHKG, RAHC-Q reaches 0.732 SI temperature MRR, 15.6\% above MAYPL; RAHC-G reaches 0.681 SI time MRR, a 6.4\% gain. RAHC-Q also leads SynMOF temperature and time at 0.71339 and 0.634 MRR, respectively, and FAIR-MOFs temperature at 0.878 SI and 0.871 OA MRR.

The G--Q comparison separates the effects of fixed reliability and contextual weighting. RAHC-Q leads most temperature and time columns, whereas RAHC-G leads MOF-SynHKG time and the solvent columns on SynMOF and FAIR-MOFs. The shared fusion mechanism can therefore use either global evidence consistency or query-dependent weighting; the results do not favor one weighting strategy for every target.

\subsection{Cross-Material Completion}
\label{sec:cross-material}

We test different chemical roles and candidate vocabularies through inorganic precursor and zeolite OSDA selection, retraining models and rebuilding rules for each task (Table~\ref{tab:non-mof}).

\begin{table}[!tbp]
\centering
\caption{Cross-material completion after task-specific retraining (MRR). SI denotes Semi-inductive; OA denotes Open-Answer Inductive. Scores are rounded to three decimals; emphasis follows unrounded rankings.}
\label{tab:non-mof}
\begin{threeparttable}
\small
\begin{fullwidthtabular}{@{\extracolsep{\fill}}l cc cc@{}}
\toprule
& \multicolumn{2}{c}{Solution Inorganic} & \multicolumn{2}{c}{ZeoSyn} \\
\cmidrule(lr){2-3}\cmidrule(l){4-5}
Method & SI & OA & SI & OA \\
\midrule
MAYPL & 0.039 & 0.035 & 0.243 & 0.201 \\
AnyBURL & 0.171 & 0.153 & 0.448 & 0.371 \\
HyperPath & 0.179 & 0.160 & 0.280 & 0.232 \\
NBFNet & 0.070 & 0.063 & 0.286 & 0.237 \\
HYPER & 0.053 & 0.047 & 0.294 & 0.243 \\
QBLP & 0.040 & 0.036 & 0.301 & 0.249 \\
RAHC-G & \second{0.334}& \second{0.299}& \best{0.465}& \best{0.385} \\
RAHC-Q& \best{0.342} & \best{0.306}& \second{0.457}& \second{0.378} \\
\bottomrule
\end{fullwidthtabular}
\end{threeparttable}
\end{table}

On Solution Inorganic, RAHC-Q reaches 0.342 SI and 0.306 OA MRR, exceeding HyperPath's 0.1792 SI MRR by 90.8\%. OA ranks 2,386 training-only precursors for 8,110 queries: 7,258 answers are in vocabulary (89.49\%); 852 receive zero reciprocal rank. On ZeoSyn, RAHC-G leads both SI and OA at 0.465 and 0.385 MRR, respectively, extending candidate-specific evidence beyond MOF conditions.

\subsection{Joint Two-Slot Completion}

We mask two slots and rank complete pairs to test cross-condition prediction. Independent uses the RAHC component scores with $\lambda_J=0$; the full variants add tuple-specific rule evidence with their selected $\lambda_J$. Backbone, checkpoint, queries, candidates, and ranking procedure are shared in this comparison. Table~\ref{tab:double-mask} also reports InGram, NS-HART, AnyBURL, and HyperPath using query-level joint MRR. Ordered and bidirectional autoregressive comparisons appear in Appendix~\ref{app:additional-results}, Table~\ref{tab:biar}.

\begin{table}[!tbp]
\centering
\caption{Joint two-slot completion (query-level joint MRR). Independent is RAHC with $\lambda_J=0$. Solv., Temp., and Method denote solvent, temperature bin, and reaction method.}
\label{tab:double-mask}
\small
\begin{adjustbox}{max width=\linewidth}
\begin{tabular}{@{}lcccc@{}}
\toprule
Method & Solv.+Temp. & Solv.+Time & Temp.+Time & Solv.+Method \\
\midrule
Independent & 0.2457 & 0.2617 & 0.4012 & 0.3907 \\
InGram & 0.2275 & 0.2484 & 0.4039 & 0.3953 \\
NS-HART & 0.2275 & 0.2610 & 0.4100 & 0.4001 \\
AnyBURL & 0.2372 & 0.2632 & 0.4075 & 0.4147 \\
HyperPath & 0.2254 & 0.2677 & 0.3901 & 0.4180 \\
RAHC-G & \best{0.2707} & \best{0.2699} & \second{0.4273} & \best{0.4197} \\
RAHC-Q & \second{0.2704} & \second{0.2685} & \best{0.4331} & \second{0.4188} \\
\bottomrule
\end{tabular}
\end{adjustbox}
\end{table}

RAHC variants lead all four tasks, including the InGram and AnyBURL comparisons. Relative to the same scorer with $\lambda_J=0$, RAHC-Q raises TemperatureBin+TimeBin joint MRR from 0.4012 to 0.4331 (8.0\%), while RAHC-G raises Solvent+ReactionMethod from 0.3907 to 0.4197 (7.4\%). RAHC-G leads three pairs and RAHC-Q leads TemperatureBin+TimeBin; both exceed Independent on every pair. These complete-tuple gains isolate the contribution of the joint-rule term under the shared scoring configuration. RAHC-G additionally exceeds both autoregressive orders and their bidirectional combination on all four comparisons in Table~\ref{tab:biar}.

\subsection{Standard Fully-Inductive KG Completion}

The materials experiments transfer to unseen query objects while retaining shared contextual entities. To test transfer across completely disjoint entity sets, we evaluate ILPC'22 Small, where each head or tail is ranked among 6,653 inference entities absent from training. Both RAHC variants use NodePieceGNN as their shared structural encoder, isolating the additional contribution of training-derived relation rules under this protocol.

\begin{table}[!tbp]
\centering
\caption{Fully-inductive link prediction on ILPC'22 Small (filtered ranking, realistic ties). H@$k$ denotes Hits@$k$; additional metrics are in Table~\ref{tab:ilpc22-full-metrics}.}
\label{tab:ilpc22-fully-inductive}
\small
\setlength{\tabcolsep}{4pt}
\begin{adjustbox}{max width=\columnwidth}
\begin{fullwidthtabular}{@{\extracolsep{\fill}}lrrrr@{}}
\toprule
Method & MRR & H@1 & H@3 & H@10 \\
\midrule
NodePiece & 0.03154 & 0.00293 & 0.01654 & 0.08012 \\
NodePieceGNN & 0.08659 & 0.03773 & 0.08925 & 0.19745 \\
NBFNet & 0.01879 & 0.00207 & 0.02395 & 0.05582 \\
AnyBURL & 0.02570 & 0.01482 & 0.02464 & 0.04118 \\
\midrule
RAHC-G & \best{0.08965} & \best{0.03894} & \best{0.09459} & \best{0.20417} \\
RAHC-Q & \second{0.08850} & \second{0.03790} & \second{0.09390} & \second{0.20262} \\
\bottomrule
\end{fullwidthtabular}
\end{adjustbox}
\end{table}

Table~\ref{tab:ilpc22-fully-inductive} reports 0.08965 MRR for RAHC-G, compared with 0.08659 for NodePieceGNN and 0.08850 for RAHC-Q. RAHC-G also leads Hits@$1$, Hits@$3$, Hits@$5$, Hits@$10$, and AMRI. Its MRR gain over NodePieceGNN is 0.00306, with a paired-bootstrap 95\% confidence interval of [0.00169, 0.00448].

Rules mined on the training graph thus provide additional ranking evidence when instantiated through observed paths among unseen inference entities. Global aggregation performs better than query-conditioned aggregation here, paralleling the task-dependent G--Q differences in materials completion. Appendix~\ref{app:ilpc22-details} reports the paired comparisons, rule coverage, and validation ablations.

\paragraph{Rule component ablation.}
We compare hard validity constraints and soft ranking rules on an element-conditioned missing-precursor validation task (Table~\ref{tab:scientific-constraint-ablation}).

\begin{table}[!tbp]
\centering
\caption{Hard- and soft-rule ablation on the element-conditioned missing-precursor validation task. H@$k$ denotes SI Hits@$k$. ``w/o hard'', ``w/o soft'', and ``w/o both'' remove hard constraints, soft rules, or both.}
\label{tab:scientific-constraint-ablation}
\small
\setlength{\tabcolsep}{3pt}
\begin{adjustbox}{max width=\columnwidth}
\begin{fullwidthtabular}{@{\extracolsep{\fill}}lccccc@{}}
\toprule
Variant & SI MRR & H@1 & H@3 & H@10 & OA MRR\\
\midrule
RAHC w/o both & 0.0491 & 0.0173 & 0.0264 & 0.0900 & 0.0466 \\
RAHC w/o hard & \second{0.4595} & \second{0.2912} & \second{0.5124} & \second{0.6983} & \second{0.4359} \\
RAHC w/o soft & 0.2585 & 0.1420 & 0.2831 & 0.4750 & 0.2452 \\
RAHC (full) & \best{0.4613} & \best{0.2972} & \best{0.5794} & \best{0.7491} & \best{0.4376} \\
\bottomrule
\end{fullwidthtabular}
\end{adjustbox}
\end{table}

On the element-conditioned missing-precursor validation task, full RAHC reaches 0.4613 SI MRR, versus 0.0491 without either rule component (Table~\ref{tab:scientific-constraint-ablation}). Soft evidence supplies most of the MRR gain: retaining it without hard rules reaches 0.4595, whereas hard rules alone reach 0.2585. Adding hard constraints raises Hits@3 from 0.5124 to 0.5794 and Hits@10 from 0.6983 to 0.7491.

\begin{table}[!tbp]
\centering
\caption{Rule-source ablation for MOF-SynHKG solvent completion (SI validation; 3,688 training-seen solvent candidates). For unseen-MOF or unseen-run queries, candidate-specific evidence comes from training-derived rules, literature-derived rules, or both; H@$k$ denotes Hits@$k$.}
\label{tab:literature-rule-ablation}
\small
\setlength{\tabcolsep}{4pt}
\begin{adjustbox}{max width=\columnwidth}
\begin{tabular}{@{}lcccc@{}}
\toprule
Model variant & MRR & H@1 & H@3 & H@10 \\
\midrule
w/o All Rules & 0.3381 & 0.2254 & 0.3740 & 0.5735 \\
w/o Literature Rules & \second{0.3726} & \second{0.2609} & \second{0.4161} & \second{0.6024} \\
w/o Training Rules & 0.3486 & 0.2400 & 0.3804 & 0.5770 \\
Training + Literature Rules & \best{0.3752} & \best{0.2624} & \best{0.4208} & \best{0.6038} \\
\bottomrule
\end{tabular}
\end{adjustbox}
\end{table}

Table~\ref{tab:literature-rule-ablation} isolates the training-derived and literature-derived rule sources within the candidate-specific evidence branch described in Section~\ref{sec:method}. Combining both sources raises SI validation MRR from 0.3726 with training rules alone to 0.3752; literature rules alone also exceed the no-rule variant (0.3486 versus 0.3381). The additional gain from literature rules in this setting is small.

\section{Conclusion}

We introduced RAHC for recovering missing conditions from scientific HKGs. Candidate-specific retrieval and dual encoding connect structural context to traceable rule evidence, while global and query-conditioned aggregation distinguish reliability from contextual relevance. Across five materials datasets, the framework improves both single-slot and joint-slot rankings; hard- and soft-rule ablations identify complementary contributions to candidate validity and preference. Training-derived relation rules also improve the shared NodePieceGNN encoder on entity-disjoint ILPC'22 Small. These evaluations support a common rule-augmented ranking framework for context-rich synthesis records and Fully-inductive KG completion.

\clearpage
\bibliographystyle{ACM-Reference-Format}
\nocite{hauptvogel2012family}
\bibliography{references}

\appendix

\section{Dataset and Task Overview}
\label{app:dataset-overview}

\begin{table}[H]
\centering
\caption{Datasets and tasks. ILPC'22 uses an entity-disjoint inference graph.}
\label{tab:dataset-overview}
\small
\setlength{\tabcolsep}{2pt}
\renewcommand{\arraystretch}{1.05}
\begin{fullwidthtabular}{@{\extracolsep{\fill}}>{\raggedright\arraybackslash}p{0.27\columnwidth}>{\raggedright\arraybackslash}p{0.16\columnwidth}>{\raggedright\arraybackslash}p{0.25\columnwidth}>{\raggedright\arraybackslash}p{0.24\columnwidth}@{}}
\toprule
Dataset & Domain & Target & Evaluation \\
\midrule
MOF-SynHKG & MOF & Solvent, temperature, time & Main HKG; multi-slot \\
SynMOF & MOF & Solvent, temperature, time & Cross-source \\
FAIR-MOFs & MOF & Solvent, temperature, time & Cross-source \\
Solution-based Inorganic & Inorganic & Precursor & Cross-material \\
ZeoSyn & Zeolite & OSDA & Cross-material \\
ILPC'22 Small & Wikidata KG & Head, tail & Fully-inductive \\
\bottomrule
\end{fullwidthtabular}
\end{table}

\section{ILPC'22 Fully-Inductive Analysis}
\label{app:ilpc22-details}

\begin{table}[h]
\centering
\caption{Additional ILPC'22 Small metrics under the protocol of Table~\ref{tab:ilpc22-fully-inductive}. AMRI is the adjusted mean rank index.}
\label{tab:ilpc22-full-metrics}
\small
\begin{fullwidthtabular}{@{\extracolsep{\fill}}lrr@{}}
\toprule
Method & Hits@5 & AMRI \\
\midrule
NodePiece & 0.03790 & 0.66999 \\
NodePieceGNN & 0.12870 & 0.69396 \\
NBFNet & 0.03946 & 0.08613 \\
AnyBURL & 0.03205 & 0.08250 \\
\midrule
RAHC-G & \best{0.13456} & \best{0.69794} \\
RAHC-Q & \second{0.13353} & \second{0.69736} \\
\bottomrule
\end{fullwidthtabular}
\end{table}

\paragraph{Benchmark and model selection.}
The ILPC'22 Small training graph contains 78,616 triples, 10,230 entities, and 48 forward relations (96 with inverses). Its inference graph contains 20,960 observed triples and 6,653 entities absent from training; validation and test contain 2,908 and 2,902 target triples. The training and inference entity sets have zero overlap. Validation and test targets are excluded from the observed inference graph. The rule branch uses 119 relation rules mined only from the training graph and instantiated through observed inference-graph paths. Validation MRR determines random seeds, checkpoints, and fusion coefficients; both RAHC variants select a coefficient of 1.0. Frozen configurations are tested once per method, yielding 34,824 saved direction-level rankings across six methods.

We performed 10,000 paired bootstrap resamples over the 2,902 original test triples. For each triple, head- and tail-prediction reciprocal ranks were averaged before resampling. Table~\ref{tab:ilpc22-bootstrap} reports the resulting MRR differences. The intervals for both RAHC variants versus NodePieceGNN exclude zero; RAHC-G also exceeds RAHC-Q under this protocol.

\begin{table}[t]
\centering
\caption{Paired-bootstrap MRR differences on ILPC'22 Small; intervals are 95\% confidence intervals over test triples.}
\label{tab:ilpc22-bootstrap}
\small
\begin{adjustbox}{max width=\columnwidth}
\begin{fullwidthtabular}{@{\extracolsep{\fill}}lrr@{}}
\toprule
Comparison & $\Delta$MRR & 95\% CI \\
\midrule
RAHC-G $-$ NodePieceGNN & +0.00306 & [0.00169, 0.00448] \\
RAHC-Q $-$ NodePieceGNN & +0.00192 & [0.00074, 0.00313] \\
RAHC-Q $-$ RAHC-G & -0.00115 & [-0.00173, -0.00065] \\
RAHC-G $-$ NBFNet & +0.07086 & [0.06534, 0.07661] \\
RAHC-Q $-$ NBFNet & +0.06971 & [0.06423, 0.07533] \\
RAHC-G $-$ AnyBURL & +0.06396 & [0.05803, 0.07016] \\
RAHC-Q $-$ AnyBURL & +0.06281 & [0.05672, 0.06904] \\
\bottomrule
\end{fullwidthtabular}
\end{adjustbox}
\end{table}

The 119 training-graph relation rules cover 38.12\% of validation prediction directions and 38.35\% of test prediction directions. On test, each prediction direction matches 1.145 rules and 15.278 inference-graph instance paths on average. Table~\ref{tab:ilpc22-rule-ablation} gives the validation ablation. Shuffling rule heads or bodies reduces RAHC-G MRR from 0.08394 to 0.08122 or 0.08123; removing inference-path evidence reaches 0.08122. Query conditioning contributes within RAHC-Q: its 0.08326 MRR decreases to 0.08165 without that component.

\begin{table}[t]
\centering
\caption{Validation ablation of ILPC'22 Small relation-rule components (MRR).}
\label{tab:ilpc22-rule-ablation}
\small
\begin{adjustbox}{max width=\columnwidth}
\begin{fullwidthtabular}{@{\extracolsep{\fill}}lr@{}}
\toprule
Configuration & Validation MRR \\
\midrule
RAHC-G, $\lambda_G=1.0$ & \best{0.08394} \\
RAHC-Q, $\lambda_Q=1.0$ & 0.08326 \\
Shuffled rule heads & 0.08122 \\
Shuffled rule bodies & 0.08123 \\
Without query conditioning & 0.08165 \\
Reliability only & 0.08174 \\
Without inference-path evidence & 0.08122 \\
\bottomrule
\end{fullwidthtabular}
\end{adjustbox}
\end{table}

\section{Additional Results}
\label{app:additional-results}

The tables below provide supplementary materials-synthesis baseline matrices and comparisons.

\begin{table}[t]
\centering
\caption{Transductive MRR across tasks.}
\label{tab:transductive}
\small
\textbf{Panel A: MOF synthesis.}\par\smallskip
\begin{adjustbox}{max width=\columnwidth}
\begin{fullwidthtabular}{@{\extracolsep{\fill}}l cccc@{}}
\toprule
& \multicolumn{3}{c}{MOF-SynHKG} & SynMOF \\
\cmidrule(lr){2-4}\cmidrule(l){5-5}
Method & Solvent & Temperature & Time & Solvent \\
\midrule
StarE & 0.41140 & 0.7256 & 0.6525 & 0.6049 \\
HAHE & 0.35007 & 0.7383 & 0.6791 & 0.6730 \\
MAYPL & 0.34307 & 0.6452 & 0.6319 & 0.5726 \\
RAHC-G & \best{0.44071} & \best{0.7904} & \best{0.7191} & \best{0.7402} \\
RAHC-Q & \second{0.42768} & \second{0.7648} & \second{0.6922} & \second{0.7360} \\
\bottomrule
\end{fullwidthtabular}
\end{adjustbox}

\smallskip
\textbf{Panel B: Non-MOF synthesis.}\par\smallskip
\begin{adjustbox}{max width=\columnwidth}
\begin{fullwidthtabular}{@{\extracolsep{\fill}}l cc@{}}
\toprule
Method & Inorganic precursor & ZeoSyn OSDA \\
\midrule
StarE & 0.3223 & 0.2975 \\
HAHE & 0.3399 & 0.3633 \\
MAYPL & 0.1416 & 0.2846 \\
RAHC-G & \best{0.3758} & \second{0.5306} \\
RAHC-Q & \second{0.3691} & \best{0.5558} \\
\bottomrule
\end{fullwidthtabular}
\end{adjustbox}
\end{table}

Table~\ref{tab:transductive} evaluates queries drawn from a shared training--test entity vocabulary. RAHC-G leads five of six tasks, while RAHC-Q leads ZeoSyn OSDA.

\begin{table}[t]
\centering
\caption{Post-hoc bidirectional autoregressive comparison for joint completion (query-level joint MRR). Independent is RAHC with $\lambda_J=0$. Solv., Temp., and Method denote solvent, temperature bin, and reaction method.}
\label{tab:biar}
\small
\begin{adjustbox}{max width=\columnwidth}
\begin{fullwidthtabular}{@{\extracolsep{\fill}}l cccc@{}}
\toprule
Method & Solv. + Temp. & Solv. + Time & Temp. + Time & Solv. + Method \\
\midrule
A$\rightarrow$B & 0.2123 & 0.2229 & 0.3738 & 0.3947 \\
B$\rightarrow$A & 0.2159 & 0.2357 & 0.3760 & 0.4035 \\
BiAR & 0.2218 & 0.2431 & 0.3931 & 0.4085 \\
Independent & 0.2457 & 0.2617 & 0.4012 & 0.3907 \\
RAHC-G & \best{0.2707} & \best{0.2699} & \second{0.4273} & \best{0.4197} \\
RAHC-Q & \second{0.2704} & \second{0.2685} & \best{0.4331} & \second{0.4188} \\
\bottomrule
\end{fullwidthtabular}
\end{adjustbox}
\end{table}

\paragraph{Exploratory three-slot completion.}
We simultaneously mask solvent, reaction-temperature bin, and reaction-time bin. Their training-only candidate sets have 3,688, 4, and 5 values, yielding 73,760 candidate tuples. All methods in Table~\ref{tab:triple-mask} use the same 1,256 test queries, observed context, ordered candidate tuples, Cartesian-product joint gold set, multi-gold filtered ranking, out-of-vocabulary (OOV) scoring, and deterministic tie-breaking. Ranking evaluates all 73,760 tuples without top-$k$ truncation. For each query, reciprocal ranks are averaged over all its joint gold tuples; a gold tuple outside the candidate space contributes zero. Query-level metrics then average over all 1,256 queries, including those with OOV answers. Joint-gold coverage is 0.95383.

\begin{table}[t]
\centering
\caption{Exploratory three-slot completion on the test set. Joint MRR and Hits@$k$ are query-level metrics over all 1,256 queries; OOV gold tuples contribute zero. All methods use the same exact filtered-ranking protocol.}
\label{tab:triple-mask}
\small
\begin{adjustbox}{max width=\columnwidth}
\begin{tabular}{@{}lcccc@{}}
\toprule
Method & Joint MRR & Hits@1 & Hits@3 & Hits@10 \\
\midrule
NS-HART adaptation & 0.13527 & 0.09315 & 0.21815 & 0.40048 \\
Independent Product & 0.13691 & 0.10669 & 0.21975 & 0.38137 \\
HyperPath adaptation & 0.13719 & 0.11146 & 0.25478 & 0.40366 \\
AnyBURL adaptation & 0.14450 & 0.11146 & 0.25000 & 0.39570 \\
RAHC-G & 0.15417 & \best{0.13462} & \best{0.27244} & 0.40224 \\
RAHC-Q & \best{0.15725} & 0.13301 & 0.27083 & \best{0.42468} \\
\bottomrule
\end{tabular}
\end{adjustbox}
\end{table}

RAHC-Q obtains the highest Joint MRR (0.15725) and Hits@10 (0.42468), while RAHC-G leads Hits@1 (0.13462) and Hits@3 (0.27244). Both variants exceed the four comparison methods in Joint MRR under this shared protocol.

\section{Traceable Solvent-Completion Example}
\label{app:solvent-case}

For test query \texttt{mayplq:00007740}, the observed context contains Zn$_4$OAc$_6$, H$_3$BBC, H$_2$bcpbd, and a temperature of $99.85\,^{\circ}\mathrm C$. The recorded solvent is DEF. Among 3,688 solvent candidates, DEF ranks 1364th without rule evidence and first with full RAHC. Six training-derived rules and one literature-derived rule match this query. A training rule relating H$_3$BBC to DEF has support 4 and confidence $4/6$. The activated literature rule is supported by Hauptvogel et al., who report using DEF to synthesize an H$_3$BBC-containing zinc coordination framework~\cite{hauptvogel2012family}. Thus, the literature supplies traceable experimental support for DEF as a candidate. The rank change reflects the combined rule evidence; this comparison does not isolate the contribution of the literature rule.

\section{Rule Construction and Reliability Details}

\label{app:rule-construction}

\begin{table}[t]
\centering
\caption{Rule taxonomy in RAHC.}
\label{tab:rule-taxonomy}
\small
\begin{adjustbox}{max width=\columnwidth}
\begin{fullwidthtabular}{@{\extracolsep{\fill}}lll@{}}
\toprule
Dimension & Types & Function \\
\midrule
Source & Data / Literature & Statistical / textual evidence \\
Strength & Hard / Soft & Validity filtering / ranking \\
Conclusion & Single-slot / Multi-slot & Value / tuple evidence \\
\bottomrule
\end{fullwidthtabular}
\end{adjustbox}
\end{table}

\paragraph{Training-data rules.}
Antecedents use generalizable features such as metal, linker, and solvent classes; functional groups; reaction methods; and temperature and time bins. Material, run, paper, and DOI identifiers are excluded to prevent memorization. On the training HKG, we enumerate antecedent combinations of length at most three and generate $A_r\Rightarrow B_r$ for each target slot. A single-slot consequent is a candidate value or class; a joint-slot consequent $B_r=(c_1,\ldots,c_k)$ is a tuple. Temperature and time bin boundaries are determined using training data only. The antecedent specifies the observed facts required to activate a rule, whereas its scope limits the material families, reaction types, and target slots to which it applies. When counterexamples appear, scope is refined and statistics are recomputed before discarding a potentially valid local pattern.

\paragraph{Literature rules from retrieval.}
The literature pipeline comprises query construction, evidence retrieval, structured extraction, and reverse verification. Retrieval queries combine the target slot, material family, and observed conditions. Relevant passages are drawn from reviews, research articles, and specialist sources. An LLM compiles only generalizable or mechanistically supported statements into $(A_r,B_r,\Omega_r)$ while retaining a source anchor. A second retrieval pass seeks independent support, scope boundaries, and counterexamples. A single experimental observation remains an evidence instance rather than automatically becoming a general rule. Chemical-name normalization, duplicate merging, and evidence consistency checks populate the supporting and counter-evidence fields. Retrieval thus constructs verifiable rules; it does not generate candidate answers directly.

\paragraph{Unified indexing.}
Training-data and literature rules undergo name, role, unit, and continuous-condition normalization; candidate generation; schema compilation; duplicate and conflict merging; hard/soft classification; reliability calibration; and indexing by target slot, scope, and antecedent.

\paragraph{Evidence-aware weighting.}
Let $N(A_r)$ count training records satisfying the antecedent and $N(A_r\cap B_r)$ those satisfying both antecedent and consequent. The data-rule statistics are
\begin{equation}
\begin{aligned}
\mathrm{supp}(r)&=N(A_r\cap B_r),\\
\mathrm{conf}(r)&=\frac{N(A_r\cap B_r)}{N(A_r)},\\
\mathrm{lift}(r)&=\frac{P(B_r\mid A_r)}{P(B_r)}.
\end{aligned}
\end{equation}
Support measures evidence volume, confidence conditional frequency, and lift enrichment over the marginal frequency. The data-rule quality vector is $Q_r^{\mathrm{data}}=f(\mathrm{supp}_r,\mathrm{conf}_r,\mathrm{lift}_r,q_r,\mathrm{stab}_r,D_r,\mathrm{ce}_r)$, where $q_r$ denotes significance, $\mathrm{stab}_r$ bootstrap stability, $D_r$ independent-source count, and $\mathrm{ce}_r$ counterexample rate. Fisher's exact test with multiple-testing correction controls chance associations; bootstrap stability measures sensitivity to training-sample perturbations. Independent supporting DOI count limits the influence of many similar experiments from one paper. For literature rules, evidence quality reflects independent sources, directness of textual support, scope consistency, and counterexamples. Source-specific quality vectors are calibrated on validation data to a raw score $R_r$. For task $t$, the reliability used by RAHC is
\begin{equation}
w_{r,t}=\frac{R_r}{\max_{r'\in\mathcal R_t}R_{r'}},
\end{equation}
where $\mathcal R_t$ is that task's formal rule catalog. Thus $w_{r,t}\in[0,1]$ is relative within a task and should not be compared across tasks. This reliability is distinct from the query-conditioned attention weight of RAHC-Q. Rule frequencies, significance, stability, source counts, and the catalog are determined after training partitioning; validation selects thresholds and calibration, and test data do not contribute to generation or weighting.

\paragraph{Completion objective.}
For a single-answer query with gold candidate $c^+$, candidate cross-entropy is
\begin{equation}
\mathcal L_{\mathrm{completion}}=-\log
\frac{\exp S(c^+\mid q)}{\sum_{c\in\mathcal C_s}\exp S(c\mid q)}.
\end{equation}
For multiple correct answers, the objective aggregates probability mass over correct candidates; joint completion applies the same objective to candidate tuples.

\section{Rule Catalogs and Reliability}
\label{app:compute}

Table~\ref{tab:rule-inventory} reports task-specific mined-rule counts and evaluation query counts. Retrieval activates the subset matching a query and candidate. For the separate HyperPath-Attr-FI baseline, the selected catalog matches a mean of 116.11 rules per test query, with coverage 1.0. HyperPath's recorded test retrieval stages take 0.517~s for Solution MissingPrecursor and 0.065~s for ZeoSyn OSDA; these measurements cover baseline rule retrieval only.

The inventory describes the catalog available to each task, not the number of rules used to score every candidate. A rule contributes only when its antecedent matches the observed query context, its scope admits the query, and its conclusion agrees with the candidate being ranked. Consequently, a larger catalog does not imply that each query receives more evidence or that the rule branch necessarily has a greater effect on MRR. The counts also differ with the target slot and the records available for mining: the MOF-SynHKG solvent catalog has 69,722 entries, whereas the corresponding temperature-bin and time-bin catalogs have 54,391 and 26,745. These are descriptive counts, not a controlled comparison of rule quantity.

\begin{table}[t]
\centering
\caption{Task-specific mined rule catalogs and evaluation query counts.}
\label{tab:rule-inventory}
\small
\begin{adjustbox}{max width=\columnwidth}
\begin{fullwidthtabular}{@{\extracolsep{\fill}}lrr@{}}
\toprule
Task & Rules & Test queries \\
\midrule
MOF-SynHKG solvent & 69,722 & 4,358 \\
MOF-SynHKG temperature bin & 54,391 & 2,462 \\
MOF-SynHKG time bin & 26,745 & 1,364 \\
Solution MissingPrecursor & 41,955 & 8,110 \\
ZeoSyn OSDA & 35,265 & 1,557 \\
HyperPath-Attr-FI baseline & 3,567 & 750 \\
\bottomrule
\end{fullwidthtabular}
\end{adjustbox}
\end{table}

The normalized-reliability summary covers each task's complete formal catalog, including zero-reliability rules. Quartiles, medians, and means use all entries. Task-wise maximum normalization makes these summaries measures of relative reliability within a task.

Table~\ref{tab:rule-reliability} shows that reliability is not uniform within a catalog. For example, the MOF solvent catalog has a median of 0.1689 and an upper quartile of 0.3408, so at least half of its rules lie at or below the median even though the task-wise maximum is 1. The joint-slot catalogs have lower medians, ranging from 0.0133 for solvent plus method to 0.0351 for temperature plus time. A joint rule must support a condition combination, making the reliability distribution particularly relevant when interpreting the joint-evidence term. These summaries describe the available evidence; they do not by themselves establish retrieval coverage, rule activation frequency, or the causal contribution of a catalog to prediction accuracy.

Because each task divides raw quality scores by its own catalog maximum, the normalized values are suitable for weighting rules within that task. They should not be read as absolute probabilities or used to rank task difficulty across datasets. In particular, the rule counts in Table~\ref{tab:rule-inventory} and the reliability summaries in Table~\ref{tab:rule-reliability} answer different questions: one records how many rules are indexed, while the other describes the evidence weights assigned to those rules. Candidate-specific retrieval and the structural branch determine how the catalog is used for an individual completion query.

\begin{table}[t]
\centering
\caption{Task-wise normalized rule reliability. All formal catalog entries, including zero-reliability rules, contribute to the summaries.}
\label{tab:rule-reliability}
\small
\begin{adjustbox}{max width=\columnwidth}
\begin{fullwidthtabular}{@{\extracolsep{\fill}}lccccc@{}}
\toprule
Task & Q1 & Median & Mean & Q3 & Max \\
\midrule
MOF solvent & 0.0799 & 0.1689 & 0.2029 & 0.3408 & 1.0000 \\
MOF temperature & 0.1232 & 0.2750 & 0.2461 & 0.3268 & 1.0000 \\
MOF time & 0.1998 & 0.2769 & 0.3313 & 0.4243 & 1.0000 \\
Inorganic precursor & 0.0308 & 0.0787 & 0.1199 & 0.1841 & 1.0000 \\
ZeoSyn OSDA & 0.1423 & 0.3320 & 0.2963 & 0.3854 & 1.0000 \\
Solvent + Temp. & 0.0046 & 0.0265 & 0.0831 & 0.0963 & 1.0000 \\
Solvent + Time & 0.0032 & 0.0237 & 0.0798 & 0.0906 & 1.0000 \\
Temp. + Time & 0.0011 & 0.0351 & 0.0939 & 0.1358 & 1.0000 \\
Solvent + Method & 0.0032 & 0.0133 & 0.0699 & 0.0638 & 1.0000 \\
\bottomrule
\end{fullwidthtabular}
\end{adjustbox}
\end{table}

\end{document}